# Biomedical Multi-hop Question Answering Using Knowledge Graph Embeddings and Language Models

Shraddha S. Mane

Persistent Systems Limited, shraddha_mane@persistent.com

Mukta A. Paliwal*

Persistent Systems Limited, paliwalmukta@gmail.com

Dattaraj J. Rao

Persistent Systems Limited, dattaraj_rao@persistent.com

Biomedical knowledge graphs (KG) are heterogenous networks consisting of biological entities as nodes and relations between them as edges. These entities and relations are extracted from millions of research papers and unified in a single resource. The goal of biomedical multi-hop question-answering over knowledge graph (KGQA) is to help biologist and scientist to get valuable insights by asking questions in natural language. Relevant answers can be found by first understanding the question and then querying the KG for right set of nodes and relationships to arrive at an answer. To model the question, language models such as RoBERTa and BioBERT are used to understand context from natural language question. One of the challenges in KGQA is missing links in the KG. Knowledge graph embeddings (KGE) help to overcome this problem by encoding nodes and edges in a dense and more efficient way.

In this paper, we use a publicly available KG called Hetionet which is an integrative network of biomedical knowledge assembled from 29 different databases of genes, compounds, diseases, and more. We have enriched this KG dataset by creating a multi-hop biomedical question-answering dataset in natural language for testing the biomedical multi-hop question-answering system and this dataset will be made available to the research community. The major contribution of this research is an integrated system that combines language models with KG embeddings to give highly relevant answers to free-form questions asked by biologists in an intuitive interface. Biomedical multi-hop question-answering system is tested on this data and results are highly encouraging.

---

* Author was a part of Persistent Systems Limited while developing this application.

## 1 INTRODUCTION

If we ask a question “Where is the CEO of Google born?” then system needs to gather the information from knowledge base such as “Sundar Pichai is the CEO of Google”, “he is born in Madurai”, “Madurai is in Tamil Nadu, India” and then answer is extracted from this knowledge – “India”. In recent NLP research, information extracted from multiple resources is effectively combined using knowledge graphs. Knowledge graph is a network of real-world entities represented by nodes and relation between them as edges. For e.g., ‘Sundar Pichai’, ‘Google’, ‘India’, ‘Tamil Nadu’ are nodes and ‘born_in’, ‘CEO_of’, ‘is_part_of’ are the edges of the graph.

KGs are used in many real-world applications such as Pinterest [1] uses bipartite graph of users and pins for the recommendation of pins. Amazon [2] uses KG to represent the products and their hierarchical relations. User information, product features are considered along with user-item bipartite graph. Self-attention as well as cross-attention mechanisms are used to achieve state of the art performance for the recommender systems. KGs are used for recommendation in Amazon prime, Amazon music. Alexa question-answering service is also based on the KGs. Search engine of Google [3] is powered by knowledge graphs as page rank algorithm. Google assistant and Google voice also make use of knowledge base to answer voice queries. Google maps uses KGs for traffic prediction, Graph Neural Network (GNN) is trained based on the traffic patterns between road segments as well as the path between source and destination. It is used to predict travel time. Facebook’s social graph is also a KG. Biomedical domain experts and scientists have started using knowledge graphs for drug discovery, drug repurposing, drug-drug interaction, drug side effects and many more. It saves a lot of time and helps to priorities the experiments to be performed in the laboratory. Recently Deepmind’s AlphaFold [4] solved very complex protein folding problem. It developed GNN based algorithm to predict 3D structure of a protein by considering sequence of Amino acids.

Over the past few years, question-answering over knowledge graph has emerged as powerful topic in research, particularly in biomedical domain. In multi-hop question answering, the system needs to perform reasoning over edges in the knowledge graph to figure out right answers. We designed a biomedical KGQA system which can answer up to 3-hop questions. Multi-hop relations are shown in figure 1.

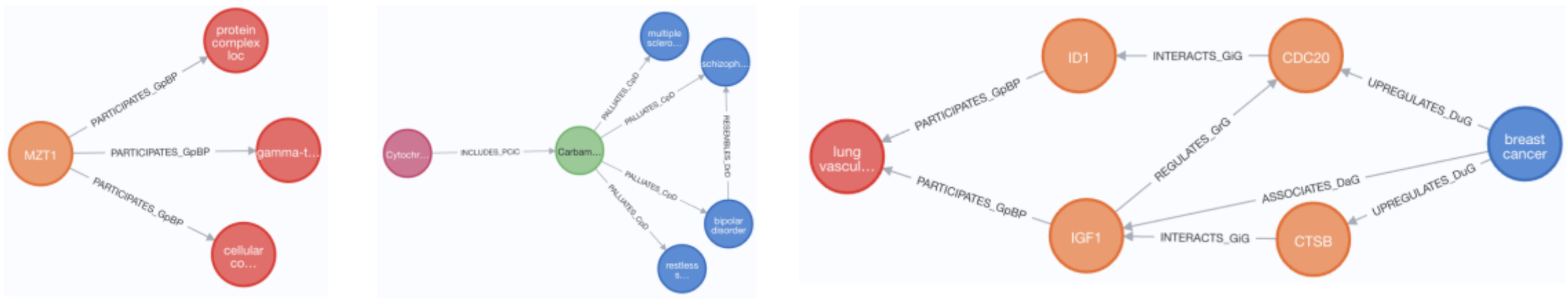

Figure 1: Multi-hop neighbourhood of the nodes. Left figure shows 1-hop relation between the MZT1 gene and 3 biological processes. 2-hop relation between Pharmacologic class and diseases is shown in the middle, path is pharmacologic class -> compound -> disease. Figure in the right shows 3-hop relation between biological process and disease, path taken is biological process <- gene <- gene <- disease

Missing links is one of the challenges in KGQA. This can be handled by using relevant external text but in many cases, it won’t be readily available. Most promising way to solve this problem is knowledge graph embeddings. As embeddings learned by KGE model extract the semantic meaning from the knowledge graph and can predict the unseen facts in the knowledge graphs. When question is asked in the natural language, it is important to understand the question and extract entities and relations mentioned in it. We used named entity recognition model to automatically extract biomedical entities from question. Once entity is extracted, we know a node of interest (i.e., head node) from a big KG and answers must be available in the neighbourhood of that node. Extracting relation from a question is nothing but understanding the context of the question. Language models such as RoBERTa [5] and BioBERT [6] are used to understand context of the questions. Question can be direct or indirect which means it can be either 1-hop, 2-hop or 3-hop. There are separate models trained for each category of questions. Question classification model identifies which class it belongs to and accordingly ML model is selected to predict the answers for given question.

Figure 2. illustrates the overview of KGQA system over biomedical data. Biomedical facts are encoded into vectors by knowledge graph embeddings and saved. Given a question – '*list all diseases that upregulate the gene which interact with gene involved in lung vasculature development*'. NER model extracts the entity from question – '*lung vasculature development*'. This is head entity i.e., node on which we are going to search for answers. Question classification module classifies this question as 3-hop question as there are 3 edges coming in path, path is *(biological process)-[gene_participates_in_biologicalprocess]-(gene)-[gene_interacts_with_gene]-(gene)-[disease_upregulates_gene]-(disease)*. Question embedding module uses RoBERTa and BioBERT and convert question into a vector which captures context of the question, so it is called as contextualized vector. Given head embedding, question embedding and embedding of all nodes in the graph, answer scoring module assigns a score to all nodes. True facts will have high score and false will have less. Top scoring nodes will be correct answers for given question. For this question 'breast cancer' and 'pancreatic cancers' are high scoring nodes. We show 10 most relevant answers for a given question.

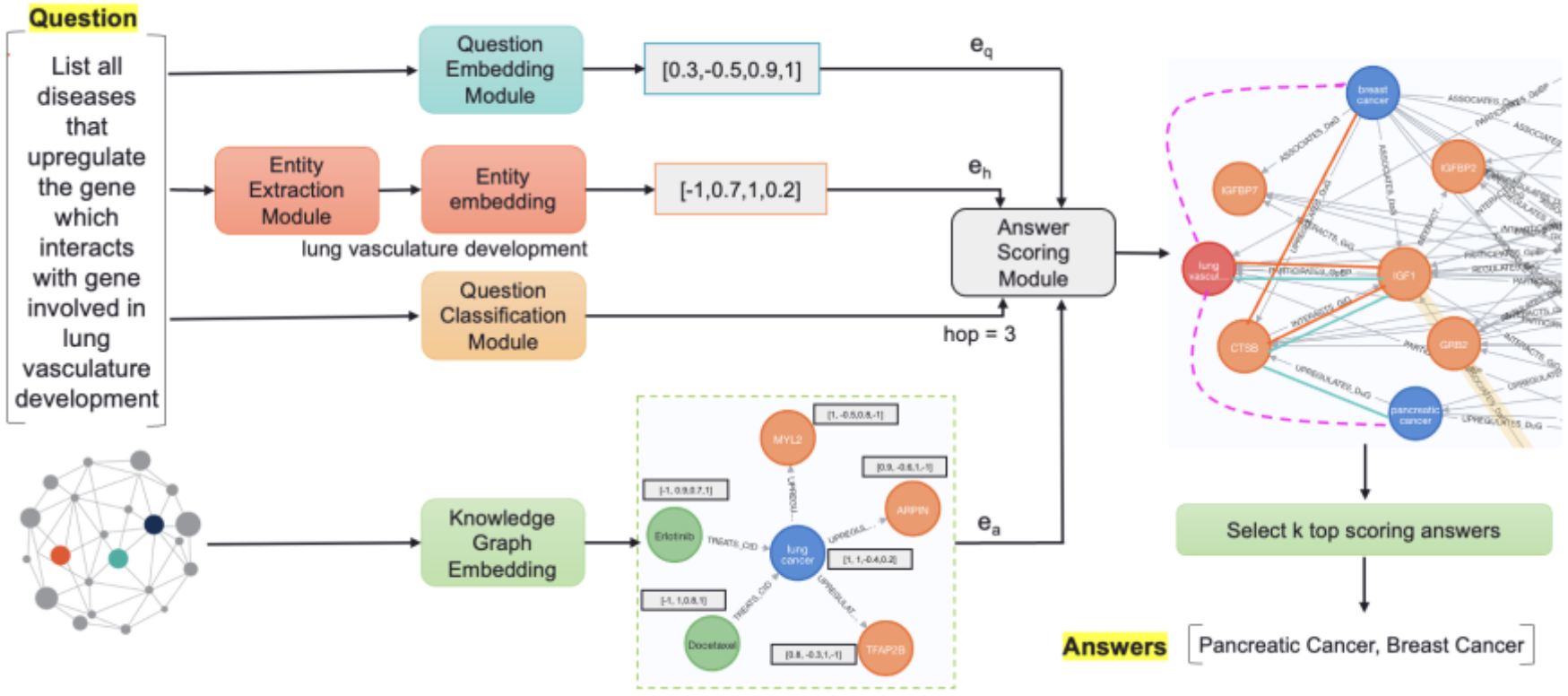

Figure 2. Overview of biomedical multi-hop question answering system.

## 2 RELATED WORK

Question answering over knowledge graphs has attracted wide attention from researchers and AI companies over the past few years. In [7], authors have presented a survey of recent research trends on Complex Question Answering over Knowledge Base. As per the paper, besides traditional methods relying on templates and rules, the general research is categorized into two main branches, namely Information Retrieval based, and Neural Semantic Parsing based. NSP based methods convert natural language query into executable language such as SPARQL. IR based methods are further divided into two categories based on their feature representation technology – feature engineering and representation learning. They have also listed several benchmark datasets involving complex questions.

The authors from the [8] have proposed QA-GNN with two key innovations: (i) relevance scoring where they use language models to find importance of KG nodes for given QA context and (ii) joint reasoning to update the representations through graph neural networks. In this paper they have used free form multiple choice data collected from professional medical board exams. Question, answer, and corresponding paragraph where answer is present are in dataset.

In [9], authors used deep cognitive reasoning network based on dual process theory in cognitive science for multi-hop KGQA. It consists of two steps – conscious and unconscious phase. Unconscious phase make use of semantic information to shortlist the candidate answers and the conscious phase uses breadth first search and Bayesian network to find accurate answers. As this approach is using sequential reasoning, it won't be able to handle missing links problem in KGQA. In [10], authors examine currently available of KGQA datasets and compare them in terms of challenges a KGQA system is facing.

Sparse knowledge graph is one of the major challenges in developing highly accurate KGQA system. In [11] authors predict the relation using embeddings. Attention based graph embedding module is used to calculate the graph embeddings by considering n-hop neighbours. Thus, learned embeddings helps to find entities of high importance from n-hop neighbourhood. As missing link problem is handled using relation prediction, KGQA achieves high F1 score. In [12] authors used attention redistribution mechanism to dynamically update question embedding and solve multi-hop KGQA problem. In [13], authors have used KG embeddings to overcome missing link challenge in KGQA. In this paper authors are providing head entity (node of interest from natural language question) along with the questions manually, which hinders the use of methodology for any practical usage.

Our approach we have used information-retrieval based method using representation learning. It uses language models to understand the context of the question and represent a question in contextualized vector form. we have extracted node of interest from natural language question using biomedical named entity recognition. It helps automating the whole process of question processing and thereby making it more useful for real life applications. Knowledge graph embeddings are used to represent all nodes and relations in dense and unique embeddings which considers similarity of nodes and relations between them. Use of KGE helps us to overcome challenging missing link problem in KGQA. Natural language question is processed using NLP to get node of interest. Given a question and node of interest, scoring function ranks all possible answers as per the confidence score to get the answers. In [7] , [10] we observed clear gap in terms of lack of complex question answering data set in biomedical field. Biomedical data published in [14] is MCQs from national medical entrance exam. And that became motivation for us to manually create question-answering data set curated from hetionet [15]. Hetionet is one of the biomedical KGs available in open source. Biomedical complex question-answering dataset is contribution from this work.

## 3 METHODOLOGY:

### 3.1 Question embedding module

This module consists of language models such as RoBERTa and BioBERT as well as linear transformation network.

#### *3.1.1Transformers*

Transformers [16] are the simple networks with self-attention mechanism. It compares each word in a sentence with every other word to check if any of them will affect one another in some way. For e.g., bank can be a bank of a river or bank for financial work. Other words present in the sentence would help to understand in which context a particular word is used. Transformers consists of two parts – encoder to read input text and decoder to produce prediction for a task for e.g., English to French translation. Training these transformers are resource-intensive task. For English French translation task this transformer achieved state of the art BLEU score after training for 3.5 days on eight GPUs. These transformers generalize well to other tasks by applying it on small training data.

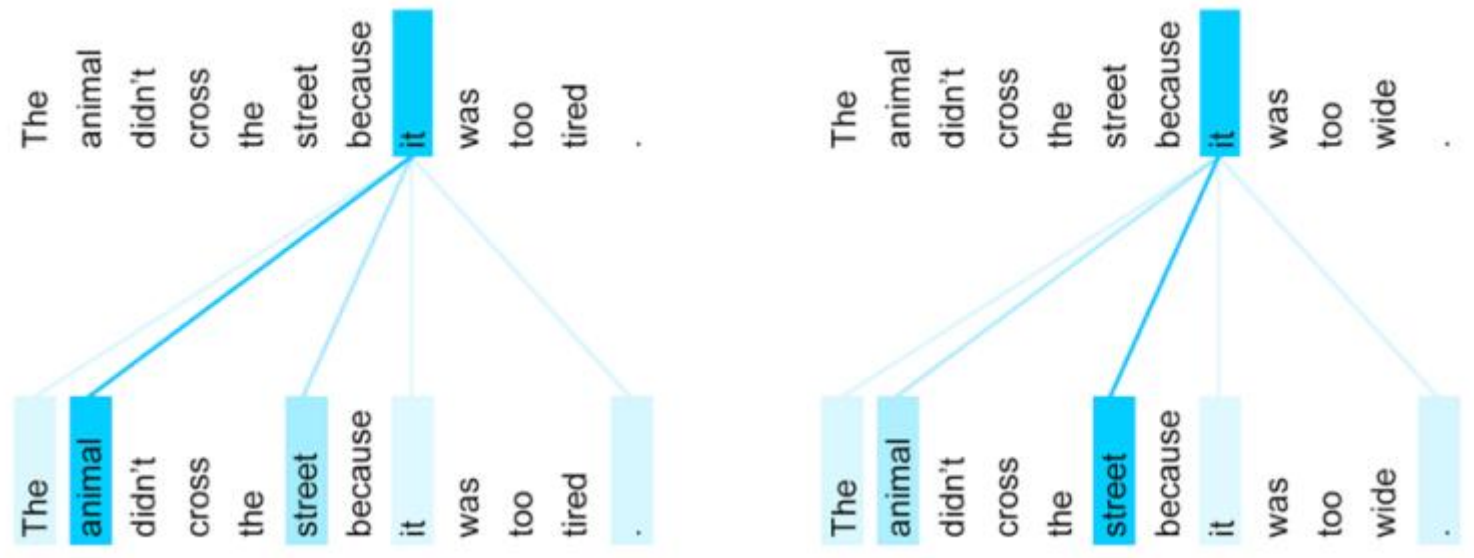

Figure 3. Transformers relate each word to every word in a sentence and score it according to the relation. Left figure shows 'it' and 'animal' relation has high score whereas in right figure 'it' and 'street' relation has high score. [17]

*3.1.2Transfer learning using language models*

Language models such as BERT [18] uses only encoder mechanism from transformers which consists of attention mechanism to learn contextual relationship between words. Transfer learning approach is used in this paper using BERT based models – RoBERTa [5] and BioBERT [6]. In KGQA system, these language models are used as a feature extractor. They convert question to a 768-dimensional feature vector. That feature vector is converted to 400-dimensional question embedding using linear transformation. Graph embeddings used in this paper are of 400 dimensions so to keep all embeddings of same size, we do this conversion. Thus, language models help to understand the question and represent in a vector form.

### 3.2 Knowledge graph embeddings

Knowledge graphs can be represented by various representations such as (subject, predicate, object) triple, adjacency matrix. However, these representations are sparse representations and can take a lot of memory. Knowledge graph embedding is a more dense and efficient representation of entities and relations present in the KG. This representation in a vector form can be utilized for different tasks. The goal of the KGE models is to encode the entities and relations such that the similarity in the embedding space approximates the similarity in the knowledge graph. A KGE is characterized by a low dimensional representation space to represent entities and relations more efficiently, scoring function to measure the similarity of triples in the embedding space and KG network and encoding models in which embeddings of nodes and relations interact with each other.

Most widely used KGE [19] models are TransE [20], TransR [21], RESCAL [22], DistMult [23], ComplEx [24] and RotatE [25]. We have used ComplEx model in our experiments. Each one of them have different scoring function to map entities and relations from network space to embedding space.

A procedure to learn representation of a graph is almost same for all models mentioned above, it is as follows:

| Knowledge graph embeddings |
| --- |

- Initialize embedding vectors for all relations and nodes in the knowledge graph to random numbers

- Repeat following steps until specified number of epochs or stopping condition (for e.g., stop training the model if there is no increment in hit@10 for 20 consecutive epochs)
- Divide given training triples (h, r, t) into 'n' small batches
- For each batch of triples, randomly generate corrupt triples such that either head, tail, or both are replaced with another entity which makes fact false. Training batch consists of true triples as well as false triples.
- Fit the model on training batches, calculate the loss. Minimize the loss by optimizing a scoring function. Update the embeddings and update model parameters.

---

After training, embeddings should extract semantic meaning from triples i.e., graph structure should be replicated in the embeddings as well.

*3.2.1ComplEx model*

ComplEx [24] is tensor decomposition model which firstly introduced complex vector space. Entities and relations are represented in complex space instead of using real valued space. Given a triple (h, r, t), complex model encodes them as ($e_h$, $e_r$, $e_t$). Such that ($e_h$, $e_r$, $e_t$)$\in \mathbb{C}^d$ , Where $\mathbb{C}^d$ is d dimensional vector in complex space. Entities are represented as combination of real and imaginary part for e.g., h = Re(h) + i Im (h). ComplEx model defines scoring function as follows

$$\phi(h, r, t) = Re(< e_h, e_r, \overline{e_t} >)$$

$$= Re(\sum_{k=1}^{d} e_h^{(k)} \, e_r^{(k)} \, \overline{e}_t^{(k)})$$

The score is more than zero for all true triples whereas it is less than zero for all negative or false triples. 'Re'` is the real part of a complex number.

## 3.3 Entity extraction and entity embedding

For a given question, extract entity from the question so that we can search on that specific node in the graph. In biomedical knowledge graph there are many entities which are not identified by spaCy's NER model. As all entities in the graph are known, we include rule-based entity recognizer using EntityRuler [26] in the existing pipeline. It boosts the accuracy of NER as entities from question are correctly extracted.

## 3.4 Question classification module

Transfer learning-based approach is leveraged here to classify a question into 1-hop, 2-hop or 3-hop class. Pretrained RoBERTa model is used as a feature extractor for a given question and then softmax layer is added on top of it to train 3-class classifier. 'AdamW' optimizer and 'CrossEntropy' loss are used while training the model.

## 3.5 Answer scoring module

This module scores all possible answers a' against given head embedding and question embedding. Entities with high score are most probable answers for given pair of question and head embedding.

$$e_{ans} = \underset{a' \in \varepsilon}{arg\,max}\, \phi(e_h, e_q, e_{a'})$$

## 4 EXPERIMENTATION DETAILS AND RESULTS:

### 4.1 Hetionet Dataset

Over 50 years of biomedical research information is combined into Hetionet [15]. It is a biomedical heterogeneous network consisting of 47031 nodes (11 types such as gene, compound, disease, pathway, biological process and more) and 2250197 relations (24 types such as BINDS_CbG, INTERACTS_GiG, PARTICIPATE_GpPW and more). Biomedical entities are represented as nodes and relations are represented as edges. This data is available in triples format - (h, r, t) along with the side information of nodes and edges.

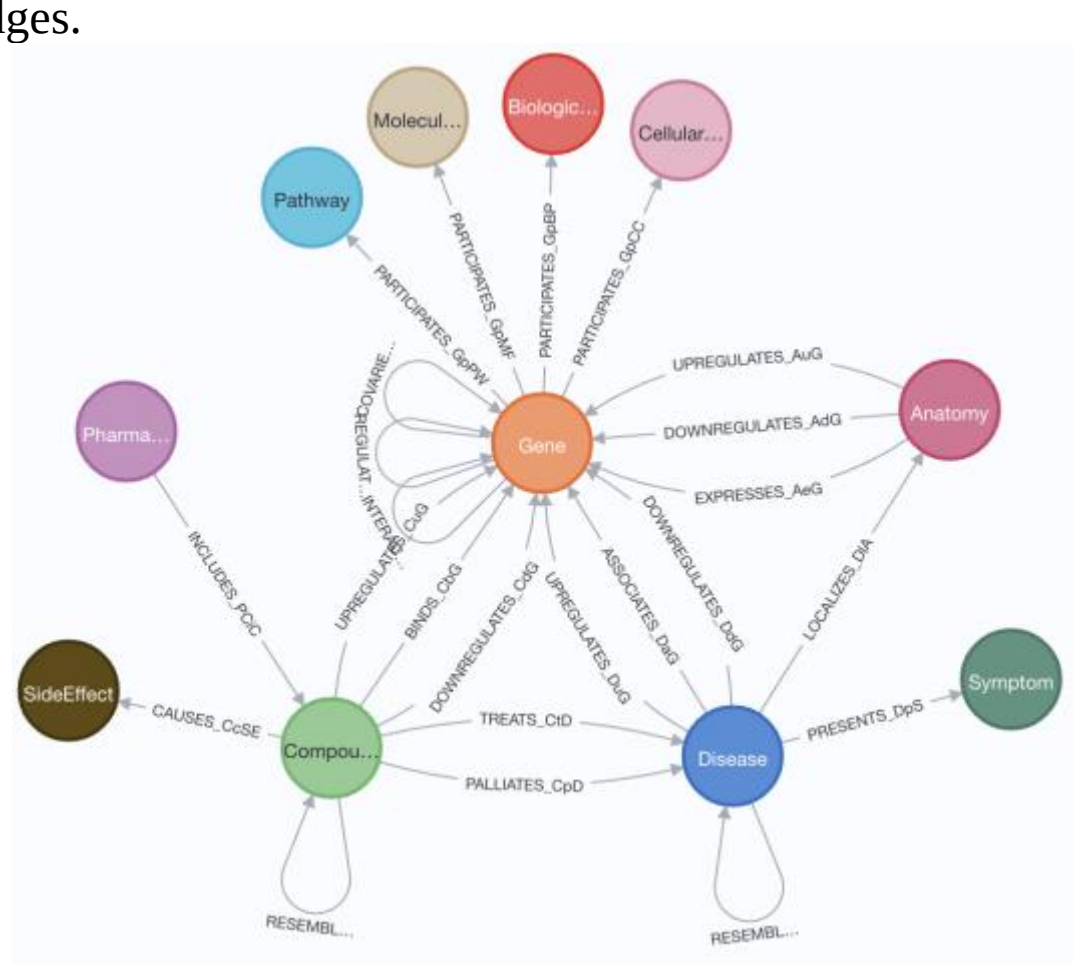

Figure 4: metagraph diagram of hetionet biomedical knowledge graph illustrates different types of nodes and connectivity between them

### 4.2 Hetionet Question-answering data

It is multi-hop QA dataset with more than 201K question-answers in the biomedical domain. It consists of 1-hop, 2-hop and 3-hop questions with one or more than one answers. Hetionet graph data is used to generate this QA dataset. We used QA data by splitting it in train-validation-test by 80:10:10. We have made hetionet question-answering dataset available for research.

Table 1: statistics of hetionet QA dataset

| | Train | Test | Validation |
|---|---|---|---|
| Hetionet QA data | 161168 | 20146 | 20146 |

### 4.3 Knowledge graph embeddings

Hetionet data available in triples – (head, relation, tail) is used to train knowledge graph embeddings. We used ‘ComplEx’ model to train KGE model. Results of ComplEx KGE model are as follows

Table 2: Complex KGE model statistics on test data

| | Adjusted Arithmetic Mean Rank | Adjusted Arithmetic Mean Rank Index | Arithmetic Mean Rank | Hits At 10 |
|---|---|---|---|---|
| ComplEx | 0.000257 | 0.9997 | 5.656 | 0.8718 |

Adjusted metrics [27] are introduced to make results comparable as rank-based metrics are affected by total number of entities in the knowledge graph. Adjusted arithmetic mean rank lies between (0,2) where lower is the better. Adjusted arithmetic mean rank index ranges between [-1,1] where higher is the better.

### 4.4 KGQA

RoBERTa and BioBERT models are used along with ComplEx KGE model. Hetionet question-answer dataset has different partitions of the dataset for 1-hop, 2-hop and 3-hop questions. We have trained separate model for each category and results are as follows:

Table 3: Results on hetionet QA dataset. Numbers mentioned in this table are hits@10 on test data

| | 1-hop | 2-hop | 3-hop |
|---|---|---|---|
| RoBETa + ComplEx | 0.7709 | 0.7705 | 0.8506 |
| BioBERT + ComplEx | 0.7421 | 0.9158 | 0.7687 |

### 4.5 Question classification

Question-answers from hetionet dataset are split into train-test-validation by ratio 80:10:10. Train and test accuracies are 1.00 and 0.99 respectively.

All the above experiments are performed on p3.2xlarge machine with 1 Tesla V100 GPU, 16GB GPU memory, 8vCPUs and network bandwidth up to 10Gbps.

## Conclusion:

In this paper, we present biomedical multi-hop question answering system using knowledge graph embeddings and language models. Knowledge graph embeddings helped to complete the incomplete knowledge graph which led to better performance of KGQA system. Biomedical KG questions answering dataset was one major challenge to measure the performance of the KGQA system. We created biomedical KGQA dataset of more than 201K question answers and made is available for research. KGQA system over biomedical knowledge graph will help researchers to prioritize the experiments to be performed in the laboratory. This will not only save a lot of time and money but also expedite the biomedical trails.